\PassOptionsToPackage{utf8}{inputenc}
\documentclass{bioinfo}

\usepackage{multirow}
\usepackage{subfigure}
\usepackage{xcolor}

\copyrightyear{2022} \pubyear{2022}

\access{Advance Access Publication Date: Day Month Year}
\appnotes{Original Paper}

\begin{document}
\firstpage{1}

\subtitle{Data and Text Mining}

\title[DxFormer]{DxFormer: A Decoupled Automatic Diagnostic System Based on Decoder-Encoder Transformer with Dense Symptom Representations}
\author[Chen \textit{et~al}.]{Wei Chen\,$^{\text{\sfb 1}}$, Cheng Zhong\,$^{\text{\sfb 1}}$, Jiajie Peng\,$^{\text{\sfb 2,3},*}$ and Zhongyu Wei\,$^{\text{\sfb 1,2},*}$} 
\address{$^{\text{\sf 1}}$School of Data Science, Fudan University, Shanghai, 200433, China and \\
$^{\text{\sf 2}}$Research Institute of automatic and Complex Systems, Fudan University, Shanghai, 200433, China and \\
$^{\text{\sf 3}}$School of Computer Science, Northwestern Polytechnical University, Xi'an, 710000, China}

\corresp{$^\ast$To whom correspondence should be addressed.}

\history{Received on XXXXX; revised on XXXXX; accepted on XXXXX}

\editor{Associate Editor: XXXXXXX}

\abstract{\textbf{Motivation:} Symptom based automatic diagnostic system queries the patient's potential symptoms through continuous interaction with the patient and makes predictions about possible diseases. A few studies use reinforcement learning (RL) to learn the optimal policy from the joint action space of symptoms and diseases. However, existing RL (or Non-RL) methods focus on disease diagnosis while ignoring the importance of symptom inquiry. Although these systems have achieved considerable diagnostic accuracy, they are still far below its performance upper bound due to few turns of interaction with patients and insufficient performance of symptom inquiry. To address this problem, we propose a new automatic diagnostic framework called DxFormer, which decouples symptom inquiry and disease diagnosis, so that these two modules can be independently optimized. The transition from symptom inquiry to disease diagnosis is parametrically determined by the \emph{stopping criteria}. In DxFormer, we treat each symptom as a token, and formalize the symptom inquiry and disease diagnosis to a language generation model and a sequence classification model respectively. We use the inverted version of Transformer, i.e., the decoder-encoder structure, to learn the representation of symptoms by jointly optimizing the reinforce reward and cross entropy loss. \\
\textbf{Results:} We conduct experiments on three real-world medical dialogue datasets, and the experimental results verify the feasibility of increasing diagnostic accuracy by improving symptom recall. Our model overcomes the shortcomings of previous RL based methods. By decoupling symptom query from the process of diagnosis, DxFormer greatly improves the symptom recall and achieves the state-of-the-art diagnostic accuracy. \\
\textbf{Availability:} Both code and data is available at \href{https://github.com/lemuria-wchen/DxFormer}{https://github.com/lemuria-wchen/DxFormer}\\
\textbf{Contact:} \href{chenwei18@fudan.edu.cn}{chenwei18@fudan.edu.cn}\\
\textbf{Supplementary information:} Supplementary data are available at \textit{Bioinformatics} online.}

\maketitle

\section{Introduction}


\begin{figure}[t]
\small
\centering
\includegraphics[width=\linewidth]{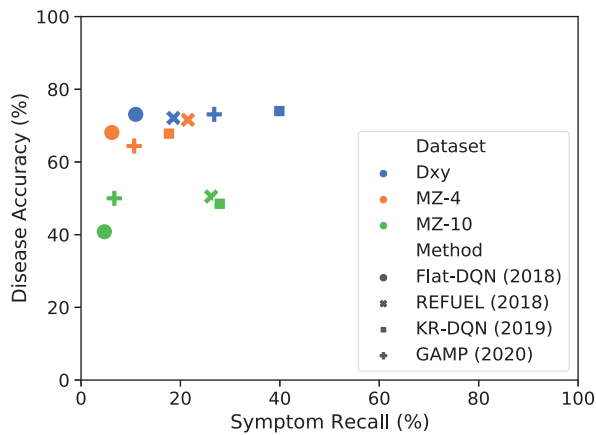}
\caption{Symptom Recall (\%) vs. Disease Accuracy (\%) of previous methods including Flat-DQN, REFUEL, KR-DQN and GAMP on three datasets on three datasets Dxy, MZ-4 and MZ-10.}
\label{fig:rec_acc}
\end{figure} 


The combination of the internet and healthcare has excellent benefits and far-reaching positive effects in improving service efficiency and promoting social equity. Automated disease diagnosis is one of the rising needs in this new healthcare model, the goal of which is to simulate the actual diagnostic process of doctors. 

The process of actual disease diagnosis can be considered as a sequence of queries and answers. Doctors choose relevant questions to ask the patient to have a better understanding of the patient's physical condition \citep{janisch2020classification}. In symptom-based automatic diagnostic system, the agent has two types of actions: one is to inquiry about a symptom, and another is to predict a disease \citep{peng2018refuel}, that is, select one of the elements from a fixed set of symptoms or diseases. The process includes several turns of \emph{symptom inquiry} and a final turn of \emph{disease diagnosis}. 

Recent years have witnessed an emerging trend of research on the task of automatic diagnosis. Considering its interactive nature, most researchers explore to model the problem by reinforcement learning (RL)~\citep{wei2018task,xu2019end,zhong2022hierarchical}. The RL setting of most studies is that, the agent chooses an action from the action space of all symptoms and diseases at each turn of interaction, and correct symptom inquiries and disease diagnoses are positively rewarded, then the policy can be learned by maximizing the expected cumulative reward.  

Although RL based approaches have made progresses for developing automatic diagnostic agents, most of them focus on disease diagnosis and lack of exploration on symptom modeling, resulting in poor symptom recall. As shown in Figure \ref{fig:rec_acc}, the symptom recall of most systems is within 30\%. In these systems, the agent often only asks the patient for one or two symptoms and rushes to make a diagnosis. In practice, however, an average of 7 to 8 symptoms are mentioned in each conversation. This results in insufficient symptom features collected by the agent, which affects the performance of the disease diagnosis.

To alleviate the above problems, we propose DxFormer, a decoupled automatic diagnostic framework, which consists of a Transformer-based decoder-encoder structure (not encoder-decoder). The decoder is for symptom inquiry, where the symptoms are treated as tokens in natural language, and the symptom inquiry is modeled as a conditional text generation task. The encoder is for disease diagnosis, where the symptoms collected in symptom inquiry are fed as the input sequence of the encoder, and the disease diagnosis is modeled as a sequence classification task. The decoder is encouraged to discover implicit symptoms, and the encoder is encouraged to make correct diagnosis, the two of which can work together in a decoupled manner and be trained simultaneously with little interference with each other. The dense representation of symptoms are learnt by optimizing the joint objectives. At runtime, the termination of symptom inquiry is determined by the stopping criterion. Specifically, the agent only switches from symptom inquiry to disease diagnosis if the confidence of the encoder on disease diagnosis reaches a certain threshold or the maximum number of turns is reached. 

To evaluate DxFormer, we conduct experiments on three real-world medical dialogue datasets: Dxy, MZ-4 and MZ-10. Experimental results verify that DxFormer greatly improves symptom recall compared with the previous methods, as well as the diagnostic accuracy. We also conduct additional ablation experiments to demonstrate the effectiveness of the components of DxFormer and further discuss the impact of the maximum number of turns and stopping criterion threshold on the model performance.

The \textbf{main contributions} of this paper can be summarized as follows: 1) We propose DxFormer, a decoupled system for automatic diagnosis based on inverted Transformer, motivated by approaching the performance upper bound of diagnostic accuracy by improving the symptom recall; 2) We discuss the impact of the maximum number of turns and stopping criterion threshold on the model performance and suggest ways to utilize DxFormer in practice; 3) Extensive experiments show that the proposed model achieves the new state-of-the-art (SOTA) results in all the three real-world datasets. 

\section{Preliminary}

\subsection{Formalization}
\label{sec:formalization}

\paragraph{MCR} ~ {In practice, real medical consultation records (MCRs) are utilized to build data-driven automatic diagnostic system \citep{wei2018task,zeng2020meddialog}. There are a large number of MCRs organized by disease categories that available in online medical communities, such as \emph{Haodafu} (\href{https://www.haodf.com/}{https://www.haodf.com/}). Generally, each MCR consists of three parts: the patient's self-provided report (i.e., self-report), multi-turn doctor-patient dialogue and the corresponding disease category. The self-report can be viewed as the first sentence in the dialogue.}

\paragraph{Symptom Attribute} ~ {Symptoms are widely present in actual doctor-patient conversations, they are the main topics discussed in medical dialogues and important basis for doctors to make diagnosis \citep{zeng2020meddialog,lin2019enhancing,10.1093/bioinformatics/btac817}. However, symptoms alone in the dialogue are less informative, additional annotations are needed to find the relationship between symptoms and patients. Generally, there are two kinds of relationships between a certain symptom and the patient: 1) \emph{Positive} (POS): the patient does have the symptom; 2) \emph{Negative} (NEG): the patient does not have the symptom. The annotator is required to find out all symptom entities mentioned in the dialogue, and identify their relationship with the patient. In this paper, we refer to this relationship as the \emph{Attribute} of the symptom.}

\paragraph{Structured MCR}  ~ {Let $\mathcal{S}$ denotes the set of all possible symptoms, $\mathcal{D}$ denotes the set of possible diseases, and $\mathcal{A}$ denotes the set of possible attributes. A structured MCR can then be denoted as: $\{(s_1, a_1), (s_2, a_2), ..., (s_n, a_n), d\}$, where $s_{i} \in \mathcal{S}$ is the $i$-th symptom that appears in the dialogue (with the self-report as the first utterance), $a_{i} \in \mathcal{A}$ is the corresponding attribute of $s_{i}$, and $d \in \mathcal{D}$ is the disease label.}

\paragraph{Explicit \& Implicit Symptoms} ~ {Generally, symptoms appearing in the self-report are regarded as explicit symptoms while the others are implicit symptoms. As a notation, we take the first $k$ symptoms as explicit symptoms, denoted as $S_{exp}=\{s_1, ..., s_k\}$, and the implicit symptoms are denoted as $S_{imp}=\{s_{k+1}, ..., s_{n}\}$. $k$ is usually small because patients usually mention only 1 or 2 symptoms in their self-reports. Implicit symptoms are unknown during inference, thus the agent needs to find as many implicit symptoms as possible to obtain a more complete symptom profile about the patient before making a diagnosis.}

\paragraph{Patient Simulator} ~ {We denote the \emph{patient simulator} as $\mathcal{P}$, and $\mathcal{P}$ can be viewed as a function whose input is any symptom and whose output is the attribute of that symptom of the patient. Note that if the symptom is not among the implicit symptoms, a \emph{Unknown} (UNK) attribute is returned.}

\paragraph{Agent} ~ {Given a patient's explicit symptoms $S_{exp}$ and their attributes, the task of the agent is to choose a symptom from $\mathcal{S}$ to interact with patient simulator $\mathcal{P}$, receive the feedback, choose the next symptom, and so on for several turns. The dialogue will terminate when the agent finally makes a diagnosis, i.e. selects a disease from $\mathcal{D}$. The goal of agent is to learn a policy, which can efficiently find implicit symptoms to obtain more complete information of the patient and finally make a correct diagnosis.}

\begin{table}
\small
\centering
\caption{The accuracy bound of SVM classifier trained on Dxy, MZ-4 and MZ-10, these classifiers differ in the features they use.}
\label{tab:bound}
\begin{tabular}{lcccc} \toprule
\textbf{Dataset} & \textbf{Acc-LB} & \textbf{Acc-UB} & \textbf{Acc-UB (P)} & \textbf{Acc-UB (N)} \\ \midrule
Dxy              & 0.644           & 0.856           & 0.808               & 0.663               \\
MZ-4             & 0.646           & 0.757           & 0.698               & 0.693               \\
MZ-10            & 0.501           & 0.706           & 0.629               &  0.618              \\ \botrule
\end{tabular}
\end{table}

\subsection{Accuracy Bound}

To explain our motivation, we discuss how to evaluate automated diagnostic systems earlier in this section. There are usually two metrics to evaluate the automatic diagnostic system, i.e., symptom recall and diagnostic accuracy.

For symptom recall (SX-Rec), it refers to the proportion of implicit symptoms that inquired by the agent. For more clarity, assuming that for a patient, the sequence of symptoms asked by the agent is $S_{agt}$, then
$${\rm SX\raisebox{0mm}{-}Rec} = \frac{\sum|S_{agt} \cap S_{imp}|}{\sum|S_{imp}|}$$
SX-Rec measures the agent's ability to find implicit symptoms, with a value between 0 and 1.

For diagnostic accuracy (DX-Acc), it refers to the proportion of correct diagnosis, and also the final metric we aim to improve. Let's further discuss the reasonable range of diagnostic accuracy. For the disease classifier, the input is the patient's symptom features finally collected by the agent, and the symptom recall determines the integrity of the symptom features. If SX-Rec equals to 0, only explicit symptoms $S_{exp}$ can be utilized as features, we call the accuracy of the system in this case the accuracy lower bound (\textbf{Acc-LB}); if SX-Rec equals to 1, both explicit symptoms and all implicit symptoms are available as features, in this way, the accuracy of the system is called the accuracy upper bound (\textbf{Acc-UB}). 

We present the accuracy bounds for the three datasets (Dxy, MZ-4 and MZ-10) in Table \ref{tab:bound} as reference. We train support vector machine \citep{cortes1995support} (SVM) classifiers using five-fold cross-validation on the training set and report the accuracy on the test set. Among them, Acc-LB means that only explicit symptoms are used as features, Acc-UB means that all symptoms are used, Acc-UB (P) means that explicit symptoms and positive implicit symptoms are used, and Acc-UB (N) means that explicit symptoms and negative implicit symptoms are used. 

The results in Table \ref{tab:bound} illustrate that the diagnostic accuracy of previous SOTA systems is far from the accuracy upper bound of SVM classifier, especially for MZ-10 (Figure \ref{fig:rec_acc}). It also suggests that both positive and negative implicit symptoms are useful for disease diagnosis. Besides, the results also give a non-rigorous reference for the agent, that is, the accuracy of a reasonable agent should be roughly between Acc-LB and Acc-UB. We mention "non-rigorous" because the performance of diagnostic accuracy of different classifiers is different. The above discussion leads to the core motivation of this paper, which is to \textbf{improve the diagnostic accuracy via improving the symptom recall}. 


\begin{figure*}[t]
\centering
\subfigure[Decoder for symptom inquiry]{
\centering
\includegraphics[width=8cm]{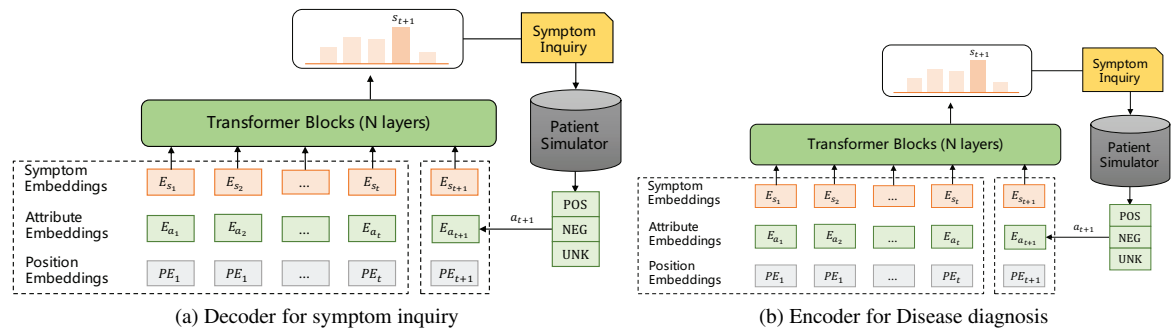}
}
\subfigure[Encoder for Disease diagnosis]{
\centering
\includegraphics[width=7cm]{figs/decoder.eps}
}
\caption{DxFormer is composed of a decoder-encoder structure, where the decoder (a) is used for symptom inquiry, and the encoder (b) is used for disease diagnosis.}
\label{fig:dxformer}
\end{figure*}

\begin{table*}
\small
\centering
\caption{Data statistics of Dxy, MZ-4 and MZ-10, the values in the form ``a/b" in the last two columns represent the \textbf{average} and \textbf{maximum} values, respectively.}
\label{tab:stat}
\begin{tabular}{lrccll} \toprule
Dataset & \# of Samples & \# of Diseases & \# of Symptoms & \# of Exp-Symptoms & \# of Imp-Symptoms \\ \midrule
Dxy \citep{xu2019end}     & 527       & 5         & 41             & 3.1 / 7    & 1.7 / 6               \\
MZ-4 \citep{wei2018task}    & 1,733      & 4         & 230           & 2.1 / 10     & 5.5 / 21              \\
MZ-10 \citep{10.1093/bioinformatics/btac817}   & 4,116      & 10        & 331         & 1.7 / 12       & 6.6 / 25             \\ \botrule
\end{tabular}
\end{table*}

\section{Method}

\subsection{Decoder for Symptom Inquiry}

For symptom inquiry, assuming that there is no limit on the number of turns, the agent can find all implicit symptoms by simply traversing the symptoms in $\mathcal{S}$. In this case, the recall equals to 1, and the accuracy can reach the upper bound. However, the size of $\mathcal{S}$, i.e., the action space of symptom inquiry, can be potentially large, this policy is undoubtedly inefficient. Fortunately, due to the apparent co-occurrence between symptoms \citep{zhong2022hierarchical}, training a more efficient agent is promising.

\subsubsection{Architecture}

In DxFormer, we analogize the process of symptom inquiry to a language model \citep{bengio2003neural}. The symptoms are regarded as tokens, and attributes of symptoms are regarded as features of tokens, then symptom inquiry can be regarded as a text generation problem. We adopt the multi-layer Transformer \citep{radford2018improving} decoder as the recurrent model, which applies a multi-headed self-attention operation over the historical symptom-attribute sequence followed by position-wise feedforward layers to produce an output distribution over target symptoms. 

\subsubsection{Input Representation}

Dense input representation are designed in this work. For each symptom, its input embedding is the sum of corresponding symptom, attribute, and position embeddings. One visual example is shown in Figure \ref{fig:dxformer}. For symptom and attribute, we utilize embedding layers to map any symptom in $\mathcal{S}$ and any attribute in $\mathcal{A}$ into dense vectors with same dimension. For position embeddings, we adopt sinusoidal position encoding used in BERT\citep{devlin2019bert}. The input sequence of symptoms is the concatenation of explicit symptoms and agent's historical asked symptoms.

\subsubsection{RL formalization}

As the decoder for symptom inquiry, our goal is to find as many implicit symptoms as possible within a given number of interactions, which is a discrete objective different from the one used in language models that maximizes the conditional likelihood. We can cast our generative model in the RL terminology as in \citep{ranzato2015sequence} due to the non-derivable property of the objective function. Our Transformer decoder can be viewed as the agent that interacts with patient simulator $\mathcal{P}$, whose parameters $\theta$ define a policy $p_{\theta}$, that results in an action that predict the next possible symptom based on known symptoms.  

\subsubsection{Reward Setting}

\paragraph{Priori Reward} ~ {A particular disease is often related to a certain group of symptoms rather than all symptoms \citep{zhong2022hierarchical}. Therefore, agent is encouraged to ask about symptoms related to the specific disease. We achieve this through the disease-symptom co-occurrence frequency matrix in training set. For each symptom in $S_{agt}$, if the frequency of co-occurrence of the symptom and the disease corresponding to the case exceeds zero, a positive reward of $+1$ will be given, otherwise a negative reward of $-1$ will be given.}

\paragraph{Ground Reward} ~ {The agent is encouraged to inquiry implicit symptoms to increase symptoms recall. For any symptom in $S_{agt}$, if the symptom is also in $S_{imp}$, a reward of $+2.5$ will be given, otherwise $-0.5$ reward will be given.}

Priori reward aims to prevent the agent from asking unrelated strange symptoms while ground reward facilitates the discovery of implicit symptoms. We will discuss the settings of these reward parameters and their impact on the results in the supplementary material.

\subsubsection{Training Objective}

The final reward of each action in $S_{agt}$ is equal to the sum of priori reward and ground reward, and the training objective of the decoder is to minimize the negative expected reward: 
\begin{equation}
    \mathcal{L}(\theta) = -\mathbf{E}_{\tau \sim p_{\theta}} [R(\tau)]
\end{equation}
where $\tau = \{r_1, r_2, ..., r_m\}$ is random reward sequence that obeys the conditional probability distribution parameterized by $p_{\theta}$, and $R(\tau) = \sum_{i=1}^{m}r_i$. 

To compute the gradient $\nabla_{\theta} L(\theta)$, we use the REINFORCE algorithm \citep{williams1992simple}, which approximates the reward function using a single Monte-Carlo sample from $p_{\theta}$ for each training example in the minibatch. Similar to traditional text generation, we initialize the parameter $\theta$ of the decoder by \textbf{language model pre-training} with maximum likelihood objectives.

\subsection{Encoder for Disease Diagnosis}

Upon the termination of symptom inquiry, we will extracted all positive and negative symptoms obtained by the agent. These symptoms, together with explicit symptoms, are used as features for disease diagnosis.

\subsubsection{Architecture}

We adopt a multi-layer Transformer encoder to encode these symptoms, and then pass through an average pooling layer and a linear layer to produce an output distribution over target diseases. As shown in Figure \ref{fig:dxformer}, the Transformer encoder in our disease classifier differs from the decoder for symptom inquiry in the following three ways: 1) The encoder adopts a bidirectional transformer, while the decoder is unidirectional; 2) Position embeddings is removed from the input representation in the encoder, since intuitively the disease classifier should be insensitive with the order of symptoms; 3) Our encoder is shallower than the decoder ($M<N$), the main reason is because the symptom inquiry is more complex and requires more parameters. It is worth noting that our encoder and decoder \textbf{share} the parameters of symptom embedding and attribute embedding.

\subsubsection{Training}

In DxFormer, the encoder is jointly trained with the decoder. Given the explicit symptoms of a patient, we obtain $S_{sdec}$ and $S_{gdec}$ by the decoder using sampling decoding and greedy decoding respectively, where $S_{sdec}$ is used to compute the REINFORCE reward, and $S_{gdec}$ is used to compute the cross-entropy loss between the predicted disease distribution and the real disease distribution. The final loss function equals to the sum of negative REINFORCE reward and the cross-entropy loss. 

\subsubsection{Stopping Criterion}

The transition from symptom inquiry to disease diagnosis is determined by the stopping criterion. During training, we specify a maximum turn $T_{max}$ for symptom inquiry, and the agent's goal is to find as many implicit symptoms as possible within $T_{max}$ turns and make correct diagnosis. However, agents do not always need to inquiry symptoms $T_{max}$ times. For some cases, the key symptom information has been collected, so it is unnecessary to continue to ask the patient to obtain some unknown or unimportant symptoms.

We take a simple but effective method to stop the symptom inquiry early. After each turn of symptom inquiry, the currently collected symptoms is fed to the encoder to compute the probability distribution over the possible diseases. The symptom inquiry will be terminated once the the probability of the predicted disease (with maximum probability) is beyond the certain threshold $\epsilon$.

More complex stop strategies may also be effective, such as combining the confidence level of the decoder~\citep{he2022bsoda}. We will leave these attempts to the future.


\begin{table*}
\small
\centering
\caption{Experimental results of DxFormer on Dxy, MZ-4 and MZ-10 dataset. The boldface values in the table are significantly better than the best baseline values at the 5\% significance level (we repeated the experiments for 10 times and conduct one-sample t-test to compute the significance).}
\label{tab:main_result}
\begin{tabular}{lccccccccc} \toprule
\multirow{2}{*}{Model} & \multicolumn{3}{c}{Dxy} & \multicolumn{3}{c}{MZ-4} & \multicolumn{3}{c}{MZ-10} \\
                       & SX-Rec    & DX-Acc   & \# Turns & SX-Rec    & DX-Acc    & \# Turns & SX-Rec    & DX-Acc    & \# Turns  \\ \midrule
\textbf{Baselines} \\ \midrule                       
Flat-DQN \citep{wei2018task}                    & 0.110 & 0.731 & 1.96   & 0.062  & 0.681  & 1.27   & 0.047  & 0.408  & 9.75    \\
REFUEL \citep{peng2018refuel}                   & 0.186 & 0.721 & 3.11   & 0.215  & 0.716  & 5.01   & 0.262  & 0.505  & 5.50    \\
KR-DQN \citep{xu2019end}                        & 0.399 & 0.740 & 5.65   & 0.177  & 0.678  & 4.61   & 0.279  & 0.485  & 5.95    \\
GAMP  \citep{xia2020generative}                 & 0.268 & 0.731 & 2.84   & 0.107  & 0.644  & 2.93   & 0.067  & 0.500  & 1.78    \\ 
BSODA  \citep{he2022bsoda}                      & -     & 0.802 & -      & -      & 0.731  & -      & -      & -      & -       \\ \midrule
DxFormer (ours)       & \textbf{0.506}  & \textbf{0.817} & 6.32     & \textbf{0.479}  & \textbf{0.743}  & 8.74     & \textbf{0.449}  & \textbf{0.633}  & 7.58
  \\  \midrule
\textbf{Ablation studies}
\\ \midrule
DxFormer-Sparse   & 0.456 &  0.808 &  7.19 &   0.396 & 0.722    & 9.37   &    0.365 &     0.619   &   8.24          \\
DxFormer-SVM       & 0.506 &  0.789 &  6.32 &   0.479 & 0.718    & 8.74   &    0.449 &     0.603   &   7.58        \\ \botrule
\end{tabular}
\end{table*}

\section{Experiments}

\subsection{Experimental Datasets}

We evaluate DxFormer on three public real-world medical dialogue datasets: Dxy, MZ-4 and MZ-10, all of which consists of a number of annotated structured MCRs described in \S~\ref{sec:formalization}. Details statistics of the datasets are listed in Table \ref{tab:stat}.

\paragraph{MZ-4} \citep{wei2018task} ~ {The first human-labeled dataset collected from the pediatric department of Baidu Muzhi Doctor (\href{https://muzhi.baidu.com/}{https://muzhi.baidu.com/}) to evaluate automatic diagnostic system. MZ-4 includes 4 diagnosed diseases: children’s bronchitis, children’s functional dyspepsia, infantile diarrhea infection, and upper respiratory infection.}

\paragraph{Dxy} \citep{xu2019end} ~ {An annotated medical dialog dataset collected from Dingxiang Doctor \footnote{\href{https://dxy.com/}{https://dxy.com/}}, a prevalent Chinese online healthcare website. Dxy includes 5 diagnosed diseases:allergic rhinitis, upper respiratory infection, pneumonia, children hand-foot-mouth disease, and pediatric diarrhea.}

\paragraph{MZ-10} \citep{10.1093/bioinformatics/btac817} ~ {A dataset with multi-level annotations expanded from MZ-4 to include 10 diseases, including typical diseases of digestive system, respiratory system and endocrine system. MZ-10 also contains more symptoms. The MZ-10 are annotated by medical students. Each dialogue is annotated twice, and the kappa coefficient of symptom labels is 92.71\%, which represents a high consistency between the two annotations.}

\subsection{Baselines}

We compare DxFormer with some state-of-the-art models for automatic diagnosis that use different techniques, including reinforcement learning, generative adversarial network, and variational autoencoder.

\paragraph{DQN} \citep{wei2018task} ~ {An agent based on the Deep Q Network (DQN) algorithm that adopts the joint action space of symptoms and diseases, where positive reward is given to the agent at the end of a success diagnosis.}

\paragraph{REFUEL} \citep{peng2018refuel} ~ {A policy based RL method with reward shaping and feature rebuilding,  where a branch to reconstruct the symptom vector is utilized to guide the policy gradient.}

\paragraph{KR-DQN} \citep{xu2019end} ~ {An improved RL method based on DQN that integrates relational refinement branches and knowledge-routed graphs to strengthen the relation between diseases and symptoms.}

\paragraph{GAMP} \citep{xia2020generative} ~ {A GAN-based policy gradient network. GAMP uses the GAN network to avoid generating randomized trials of symptom, and add mutual information to encourage the model to select the most discriminative symptoms.}

\paragraph{BSODA} \citep{he2022bsoda} ~ {An non-RL bipartite framework that uses an information-theoretic reward to collect symptoms, and a multimodal variational autoencoder (MVAE) model is used for disease prediction with a two-step sampling strategy.}

We use the open source implementation\footnote{\href{https://github.com/Guardianzc/DISCOpen-MedBox-DialoDiagnosis}{https://github.com/Guardianzc/DISCOpen-MedBox-DialoDiagnosis}} for DQN, REFUEL, KR-DQN and GAMP, since none of these papers provide official codes, and symptom recall is also not reported in most papers. 

\subsection{Model Configuration}

DxFormer is composed of a 4-layer decoder and a 1-layer encoder. For Dxy, the embedding and hidden size is set to 128, and feed-forward size is set to 256; For MZ-4 and MZ-10, the embedding and hidden size is set to 512, and feed-forward size is set to 1024. We use Adam optimizer \citep{kingma:adam} with a learning rate of $3 \times 10^{-4} $ for maximum-likelihood pre-training, and learning rate of $1 \times 10^{-4} $ for RL training. All our experiments are performed on 4 Nvidia Tesla V100 32G GPUs. Following the conventional setting, all baseline models as well as our DxFormer specify the maximum number of turns for symptom inquiry to 10 during inference, the threshold $\epsilon$ is set to 0.99. It is worth noting that during training, the maximum number of turns of DxFormer is set to 40.

\subsection{Main Findings}

In Table \ref{tab:main_result}, we report the performance of DxFormer as well as the baseline models. DxFormer shows impressive performance. On the three real-world datasets, DxFormer can find about 45$\sim$52\% of implicit symptoms in 6$\sim$8 turns, and reach about 64$\sim$84\% of diagnostic accuracy. Compared to best baseline models, DxFormer greatly improves the symptom recall, with an absolute improvement of nearly 12$\sim$27\%. Besides, the diagnostic accuracy also surpasses all previous models. In particular, on MZ-10, DxFormer improves the accuracy by about 14 absolute percentages over the best baseline.

Considering the performance of these baseline models on the MZ-10 dataset, the diagnostic accuracy is only on par with the lower bound of the SVM classifier (see in Table \ref{tab:bound}), which suggests that these systems approximately degenerate into disease classifiers that are very weak at finding implicit symptoms. This illustrates the advantages of DxFormer over other systems when faced with more diseases and symptoms. 

It is worth noting that in DxFormer, agent interacts with patient simulator more rounds (\# Turns) than most systems. It is not surprising since enough interaction turns are the guarantee of high symptom recall. As we mentioned in the introduction, the number of interactions is not controllable in traditional RL-based methods. It is difficult for us to compare the model performance within the same number of turns. Nevertheless, from a clinical point of view, one would expect that asking 6-9 symptoms on average to improve diagnostic accuracy is acceptable.


\begin{figure*}
\centering
\subfigure[MZ-4 dataset]{
\centering
\includegraphics[width=8cm]{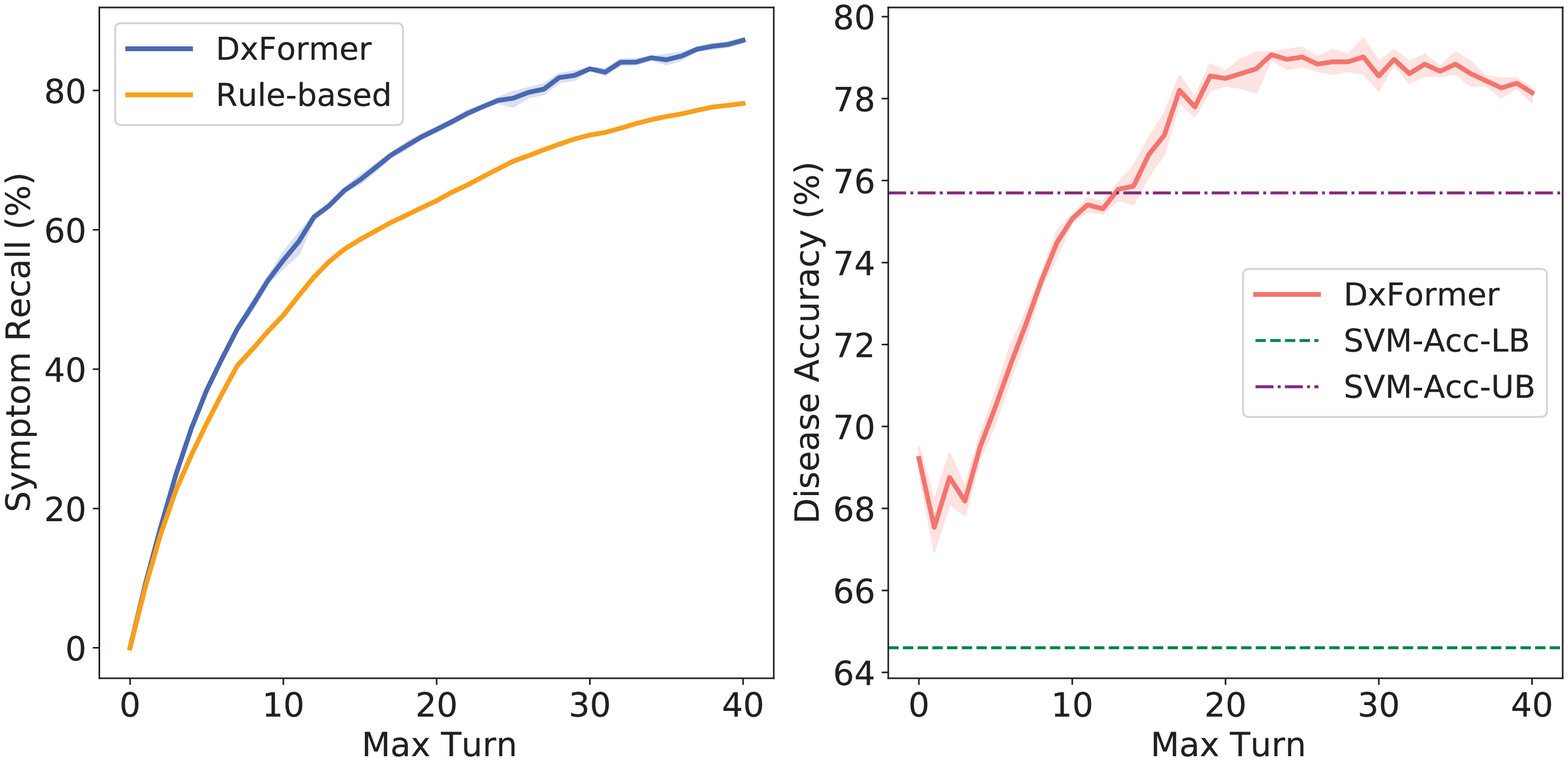}
}
\subfigure[MZ-10 dataset]{
\centering
\includegraphics[width=8cm]{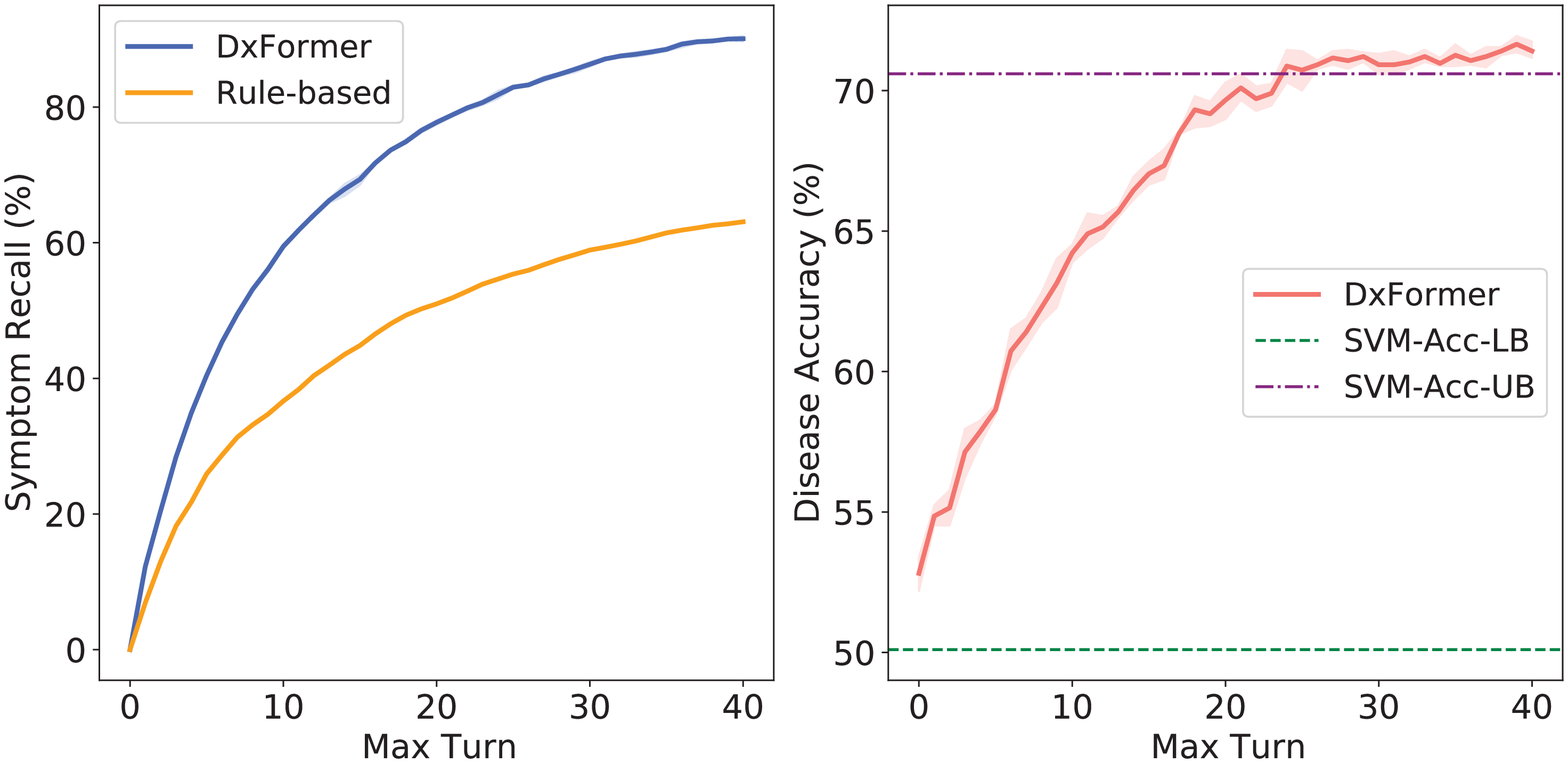}
}
\caption{The effect of max turns on the Symptom Recall and Diagnostic Accuracy on MZ-4 and MZ-10 dataset.}
\label{fig:mz4_mz10}
\end{figure*}


\begin{figure}[t]
\small
\centering
\includegraphics[width=\linewidth]{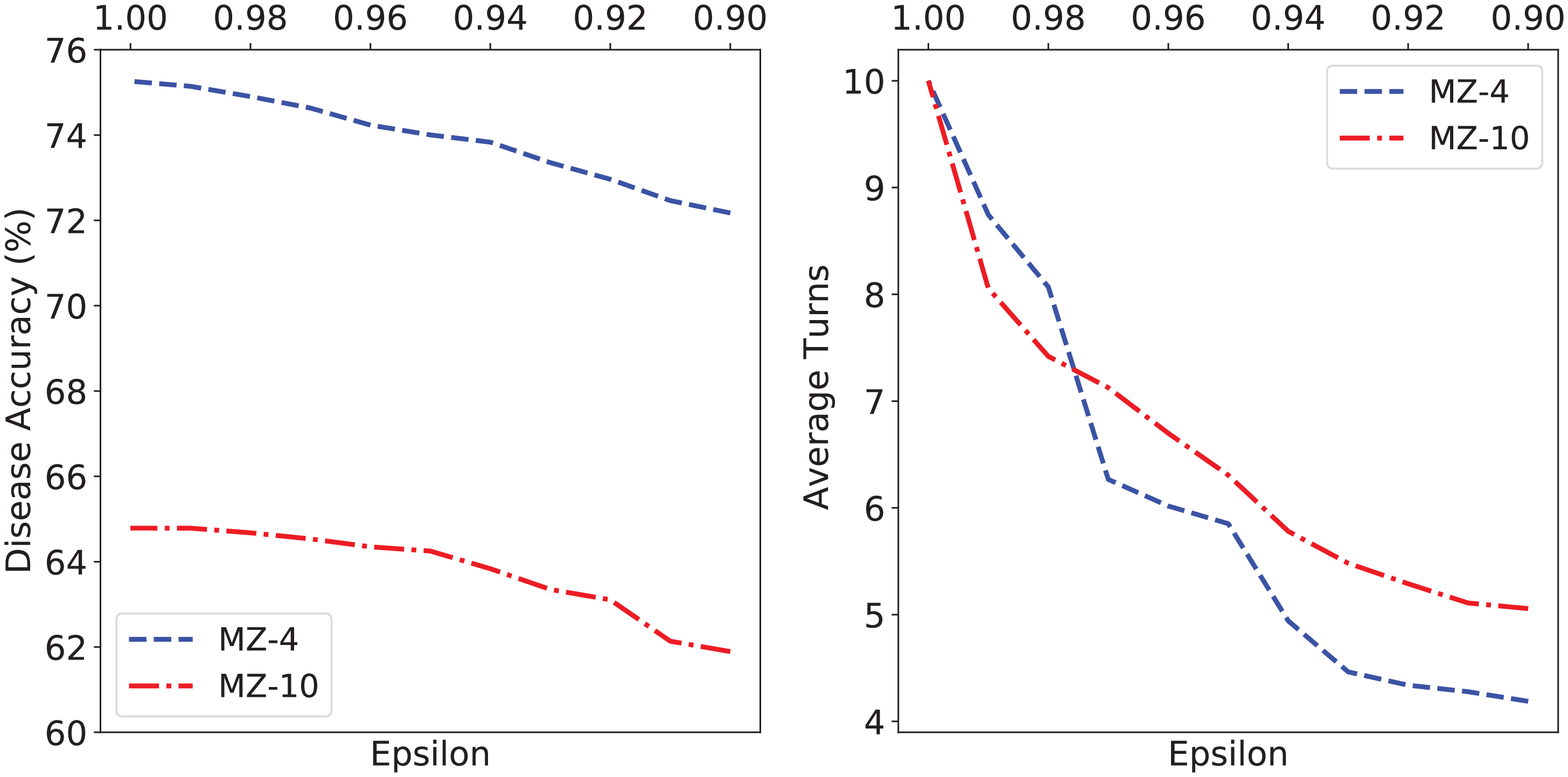}
\caption{The effect of threshold on MZ-4 and MZ-10 dataset.}
\label{fig:epsilon}
\end{figure}


\subsection{Effect of Max Number of Turns}

In figure \ref{fig:mz4_mz10}, we visualize the effect of max number of turns, i.e., $T_{max}$, on model performance on the MZ-4 and MZ-10 datasets. It can be seen that both symptom recall and diagnostic accuracy increase with the increase of the maximum number of allowed interactions. This confirms the core motivation of this paper, that \textbf{it is feasible to improve the diagnostic accuracy by increasing the symptom recall}. It can also be found that when symptom recall exceeds about 70\%, diagnostic accuracy gradually converges to the upper bound, which is higher than the upper bound of SVM classifier (purple dashed line in the figure). On the MZ-4 and MZ-10 datasets, when the agent is allowed to interact with the patient more than 20 times, the diagnostic accuracy can be as high as about 78\% and 70\%, respectively, which is much higher than the current SOTA results. 

Besides, we also create a strong rule-based agent for symptom inquiry, that is, based on the symptom co-occurrence probability matrix, to find the symptom with the highest correlation with the collected symptom set each time. The yellow line in the figure \ref{fig:mz4_mz10} shows the results of the method. It can be seen that rule-based agent performs worse than DxFormer, especially on the MZ-10 dataset. When the number of symptoms increases (from MZ-4 to MZ-10), the dense representation of symptoms shows advantages, as the symptom recall of DxFormer dose not decrease.

\subsection{Effect of Stopping Criterion Threshold}
\label{sec:balance}

We further explore the impact of the threshold value in stopping criterion on model performance. We fix the maximum number of turns to 10 and set different thresholds to observe the model performance. The results are visualized in Figure \ref{fig:epsilon}. We have two findings. First, both diagnostic accuracy and average turns decrease as the threshold value decreases from 1.0 to 0.9, this shows that, as the number of turns for symptom inquiry increases, the collected symptoms increase, and the certainty of disease classifier also increases. Secondly, symptom recall declines slowly at the beginning. This suggests that it is possible to choose an appropriate threshold that reduces the average number of turns with little loss of diagnostic accuracy.

In DxFormer, $T_{max}$ and $\epsilon$ together determine the balance of accuracy and efficiency. In practice, we recommend to first choose the $T_{max}$ that allows diagnostic accuracy to converge to the upper bound, and then select an appropriate threshold to achieve acceptable accuracy and efficiency. 

\subsection{Ablation Studies}

To verify the effectiveness of each component in DxFormer, we conduct some ablation experiments. \textbf{DxFormer-Sparse} is an agent that exactly the same as DxFormer except for the input presentation. DxFormer-Sparse uses the one-hot representation of symptom and attribute and the concatenation of the one-hot vectors are fed as the input; \textbf{DxFormer-SVM} is an agent that uses the same decoder of DxFormer but replaces the encoder with a SVM classifier. 

From the ablation analysis results in Table \ref{tab:main_result}, both DxFormer-Sparse and DxFormer-SVM perform worse than DxFormer, which shows that the dense representation of symptoms is effective. Notably, the two variants of DxFormer can still beat SOTAs, illustrating the effectiveness of our decoupled framework. In fact, in our early attempts, models like RNN-MLP, LSTM-MLP can also work well, although not as well as DxFormer. The decoder-encoder framework and stopping criterion together contributes to the excellent performance of DxFormer. 

\section{Related Work}

\paragraph{Automatic Disease Diagnosis} ~ {Machine learning based method plays an important role in solving medical prediction, such as depression prediction \citep{aekwarangkoon2022associated} and epidemic prediction \citep{kim2021prediction}. Automatic disease diagnosis is an important part of medical prediction. Deep reinforcement learning \citep{mnih2015human,silver2016mastering} has 
been applied for automatic diagnosis \citep{tang2016inquire,kao2018context}. \citep{peng2018refuel} proposed reward shaping and feature rebuilding method for fast disease diagnosis. However, their data used is simulated that cannot reflect the situation of the real diagnosis. For the medical dialogue system for automatic diagnosis, \citep{wei2018task} annotated the first medical dataset for dialogue system and use a Deep Q-network (DQN) to collect additional symptoms via conversation with patients. \citep{xu2019end} released another medical dataset for the dialogue system and introduce prior knowledge to improve the diagnosis accuracy. \citep{zhong2022hierarchical} propose a hierarchical reinforcement learning framework based on master-worker structure for simulating real medical consultation. There are also some related studies based on non-RL framework \citep{xia2020generative,he2022bsoda}. Diaformer \citep{chen2022diaformer} is a contemporary work with our DxFormer, which is similar to our ideas, but with different motivations. We do not compare the results because the number of turns is set differently.}

\paragraph{RL based Text Generation} ~ {RL is also a popular alternative in text generation, especially when the training objective is not the traditional maximum likelihood. \citep{ranzato2015sequence} Use REINFORCE algorithm to maximize the BLEU of the generated sequence to solve the exposure bias problem of the traditional seq2seq model. \citep{li-etal-2016-deep} use policy gradient algorithm to maximize the mutual information of generated response in dialogue system. \citep{rennie2017self} propose self-critical sequence training (SCST), which utilizes the output of its own test-time inference algorithm to normalize the rewards it experiences in image captioning. }

\section{Conclusions, Limitations and Future Work}

In this work,  we propose an decoupled system for automatic diagnosis called DxFormer, which uses Transformer's decoder and encoder for symptom query and disease diagnosis with dense symptom representations.  Symptom query and disease diagnosis are formalized into conditional text generation and sequence classification. The stopping criterion controls the transition from symptom query to disease diagnosis to parametrically control the number of interaction turns. Extensive experiments show that DxFormer can greatly improve the symptom recall rate and diagnostic accuracy.

Although promising results have been obtained, the system still has some limitations: 1) In this paper, the stopping criterion is simple and crude, and its reliability and scalability should be discussed; 2) In practice, agents should be allowed to ask multiple symptoms at one time rather than a single symptom to improve efficiency; 3) It is not enough to use only the symptoms for diagnosis. In the actual situation, more factors need to be considered, including medical examination, past medical history, surrounding environment, etc.

In future work, we hope to further explore automatic disease diagnosis based on more features. Besides, we will also try to improve the model from the perspective of biological interpretability.

\section{Funding}

This work is partially supported by Natural Science Foundation of China (No.71991471, No.6217020551), Science and Technology Commission of Shanghai Municipality Grant (No.20dz1200600, 21QA1400600) and Zhejiang Lab (No. 2019KD0AD01).

\bibliographystyle{natbib}
\bibliography{document}
\end{document}